\pdfoutput=1

\documentclass[11pt]{article}

\usepackage{acl}

\usepackage{times}
\usepackage{graphicx}
\usepackage{latexsym}
\usepackage{multirow}
\usepackage{amsfonts}
\usepackage[T1]{fontenc}

\usepackage[utf8]{inputenc}

\usepackage{microtype}

\title{EmoCaps: Emotion Capsule based Model for\\ 
	Conversational Emotion Recognition}

\author{Zaijing Li\textsuperscript{1}, \ Fengxiao Tang\textsuperscript{1}*, \  Ming Zhao\textsuperscript{1}*,\ Yusen Zhu\textsuperscript{2}
	\\ \textsuperscript{1}School of Computer Science and Engineering, Central South University, Changsha, China \\ \textsuperscript{2}School of Mathematics, Hunan University, Changsha, China \\ \texttt{\{lizaijing,tangfengxiao,meanzhao\}@csu.edu.cn} \\ \texttt{zhu\_yusen@163.com}}

\begin{document}
\maketitle
\begin{abstract}
	Emotion recognition in conversation (ERC) aims to analyze the speaker's state and identify their emotion in the conversation. Recent works in ERC focus on context modeling but ignore the representation of contextual emotional tendency. In order to extract multi-modal information and the emotional tendency of the utterance effectively, we propose a new structure named Emoformer to extract multi-modal emotion vectors from different modalities and fuse them with sentence vector to be an emotion capsule. Furthermore, we design an end-to-end ERC model called EmoCaps, which extracts emotion vectors through the Emoformer structure and obtain the emotion classification results from a context analysis model. Through the experiments with two benchmark datasets, our model shows better performance than the existing state-of-the-art models.
\end{abstract}
\section{Introduction}
Emotion recognition in conversation (ERC) is a work that recognizes the speaker's emotion and its influencing factors in the process of conversation. Nowadays, social media such as Facebook and Twitter generate a large amount of dialogue data with various modalities of textual, audio, and video all the time. The study of speaker emotional tendency has huge potential value in the fields of public opinion analysis, shopping, and consumption. Therefore, conversation emotion recognition has attracted more and more attention from researchers and companies. 

In ERC, existing research mainly focuses on the way of contextual information modeling (Majumder et al., 2019; Ghosal et al., 2019). However, These models have some shortcomings due to their inability to better extract the grammatical and semantic information of the utterance. Recent studies (Yuzhao Mao et al., 2020; Weizhou Shen et al., 2020) have introduced the transformer structure into the utterance feature extraction to solve the above problems. Li et al. (2021) proposed a new expression vector, "emotion vector" for ERC, which is obtained by mapping from sentence vector, but only for textual modality. Meanwhile, existing studies (Song et al., 2004; Dellaert et al., 1996; Amir, 1998) have shown that only textual information is not enough for emotional presentation, the tone and intonation reflect the speaker’s emotions to a certain extent, and the facial expressions also express the inner feelings of the speaker in most cases. As shown in Figure 1, different modalities contain different information, and all are slightly flawed, so multi-modal based information can better identify the speaker's emotion than a single modality in ERC. 
\begin{figure}[t]
	\centering
	\includegraphics[width=0.9\columnwidth]{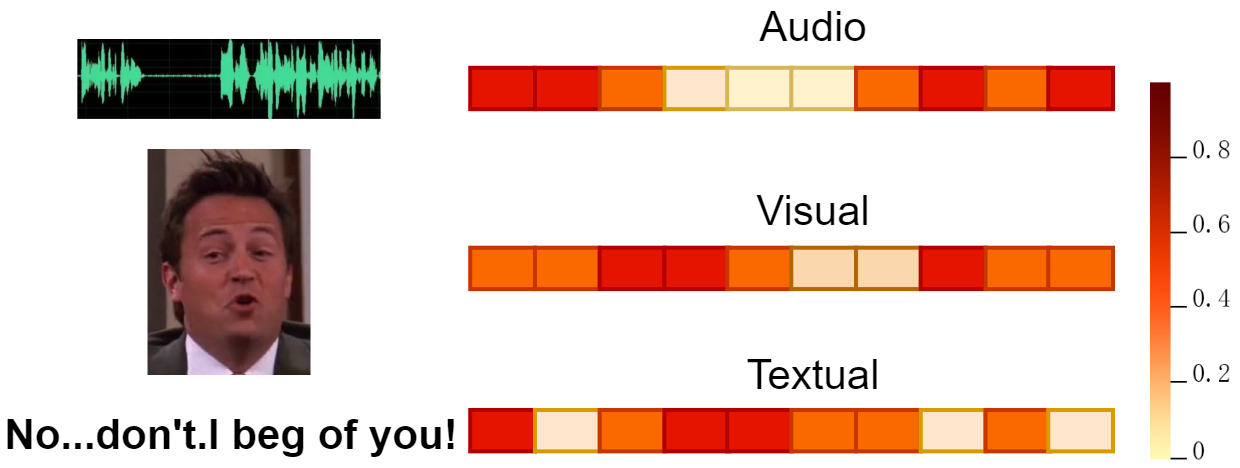} 
	\caption{Visualization of the heatmap for an utterance in a conversation, with three modalities.}
	\label{fig0}
\end{figure}

In order to identify the speaker’s emotion in conversation effectively, it is necessary to obtain good utterance features. Also we can't ignore the role of the utterance’s emotional tendency. As shown in Figure 2, the emotional tendency of the utterance itself is like an "offset vector", which makes the neutral utterance have an "emotional direction". For single-sentence emotion classification, emotional tendency is consistent with the results of emotion recognition, while in ERC, the influence of the context may cause the emotional tendency to be inconsistent with the result of emotion recognition. However, emotional tendency can provide features for the model so that model can "understand" the reason for emotional reversal.

So, we propose a new multi-modal emotional tendency extraction method called Emoformer, which is a Transformer-based model but doesn't include the decoder part. As shown in Figure 3, Emoformer extracts the emotional tendency, i.e., emotion vector, from the modal features through the multi-head self-attention layer and feed-forward layer. More details we will analysis in Section 3.
\begin{figure}[t]
	\centering
	\includegraphics[width=0.9\columnwidth]{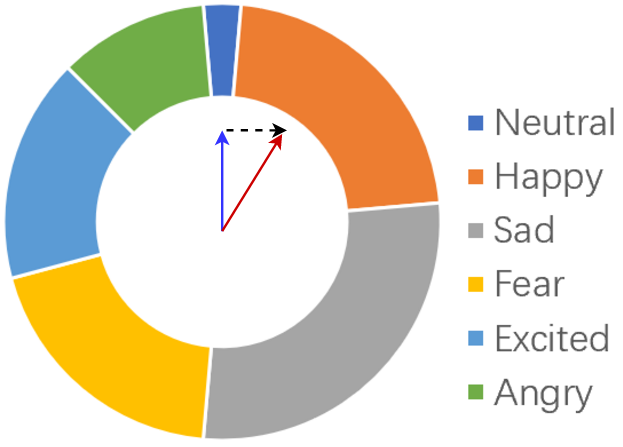} 
	\caption{A map for emotion classification. The dashed arrow represent the offset vector, which is added to the neutral vector to obtain the vector with an emotional direction.}
	\label{fig1}
\end{figure}

Based on the Emoformer, we further propose an end-to-end ERC model to classify the emotion based on multi-modal information, named EmoCaps. Specifically, we employ the Emoformer structure to extract emotion vectors of textual, audio, and visual features. Then, we merge the emotion vectors of the three modalities with the sentence vector to an emotion capsule. Finally, we employ a context analysis model to get the final result of the emotion classification.

In general, the contributions of this paper are as follows:

\begin{itemize}
	
	\item We innovatively introduce the concept of emotion vectors to multi-modal emotion recognition and propose a new emotion feature extraction structure, Emoformer, which is used to jointly extract emotion vectors of three modalities and merge them with sentence vector to the emotion capsule. 
	\item Based on Emoformer, we further propose an end-to-end emotion recognition model named EmoCaps to identify the emotion from multi-modal conversation.  
	\item Our model and the existing state-of-the-art model are tested on MELD and IEMOCAP datasets. The test results show that our model has the best performance both in multi-modality and text-modality.
\end{itemize}

The rest of the paper is organized as follows: Section 2 discusses related works; Section 3 introduces the proposed EmoCaps model in detail; Section 4 and 5 present the experiment setups on two benchmark datasets and the analysis of experiment results; Finally, Section 6 concludes the paper.
\begin{figure}[t]
	\centering
	\includegraphics[width=0.8\columnwidth]{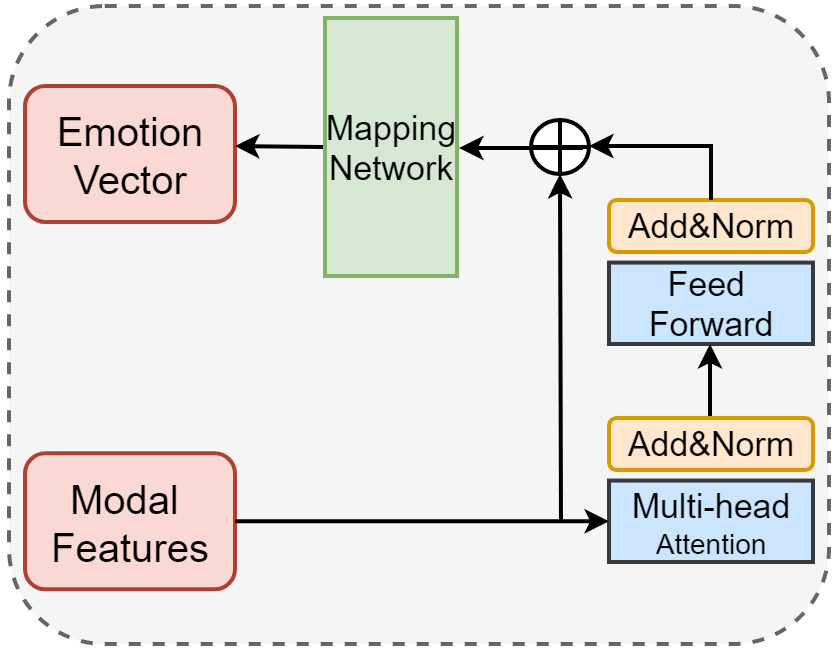} 
	\caption{Schematic diagram of Emoformer. The mapping network consists of 5 fully connected layers.}
	\label{fig2}
\end{figure}
\section{Related Work}
\subsection{Emotion Recognition in Conversation}
Poria et al. (2017) use Biredectional LSTM (Hochreiter and Schmidhuber 1997) in ERC, which builds context information without differentiating among the speakers. ICON (Hazarika et al., 2018b) is an extension of CMN (Hazarika et al.,2018), which contains another GRU structure to connect the output in the CMN model to distinguish the speaker relationship. Majumder et al. (2019) use three GRUs to obtain context information and update the speaker status. Ghosal et al. (2019) construct a conversation into a graph, then use a graph convolutional neural network to convert the emotion classification task of the conversation into a node classification problem of the graph. Ghosal et al. (2020) use common sense knowledge to learn the interaction of interlocutors. Shen et al. (2021) design a directed acyclic neural network for encoding the utterances. Hu et al. (2021) propose the DialogueCRN to fully understand the conversational context from a cognitive perspective. 
\subsection{Multi-modal Emotion Recognition}
Zadeh et al. (2017) propose the TFN model, which is a multi-modal method using the tensor outer product. Liang et al. (2018) propose the model which use a multi-level attention mechanism to extract different modal interaction information. Cai et al. (2019) propose a hierarchical fusion model to model graphic information for irony recognition. But the above models are not applied in ERC. Hazarika et al. (2018b) propose the CMN model in ERC, which uses a GRU structure to store multi-modal data information and considers the role of contextual information in conversation emotion recognition. Jingwen Hu et al. (2021) propose the MMGCN model, which is a graph convolutional neural network model based on a multi-modal hybrid approach. 
\subsection{Transformer Models}
Inspired by the self-attention mechanism (Bengio et al. 2014), the Transformer is proposed for computing representations and efficiently obtaining long-distance contextual information without using sequence (Vaswani et al. 2017), which has achieved great success in the field of computer vision and audio processing (Tianyang Lin et al. 2021). Devlin et al. (2019) use the Transformer structure to train a large-scale general-purpose text corpus to obtain a language model with syntactic and semantic information. By employing a transformer-based pretraining model, Hazarika et al. (2020) transfer the context-level weight of the generated conversation model to the conversation emotion recognition model. Yuzhao Mao et al. (2020) use Transformer to explore differentiated emotional behaviors from the perspective of within and between models. Weizhou Shen et al. (2020) use the XLNet model for conversation emotion recognition (DialogXL) to obtain longer-term contextual information. The above-mentioned algorithms use a transformer-based structure but they are not applied for multi-modal models.
\section{Methodology}
\begin{figure*}[t]
	\centering
	\includegraphics[width=0.8\textwidth]{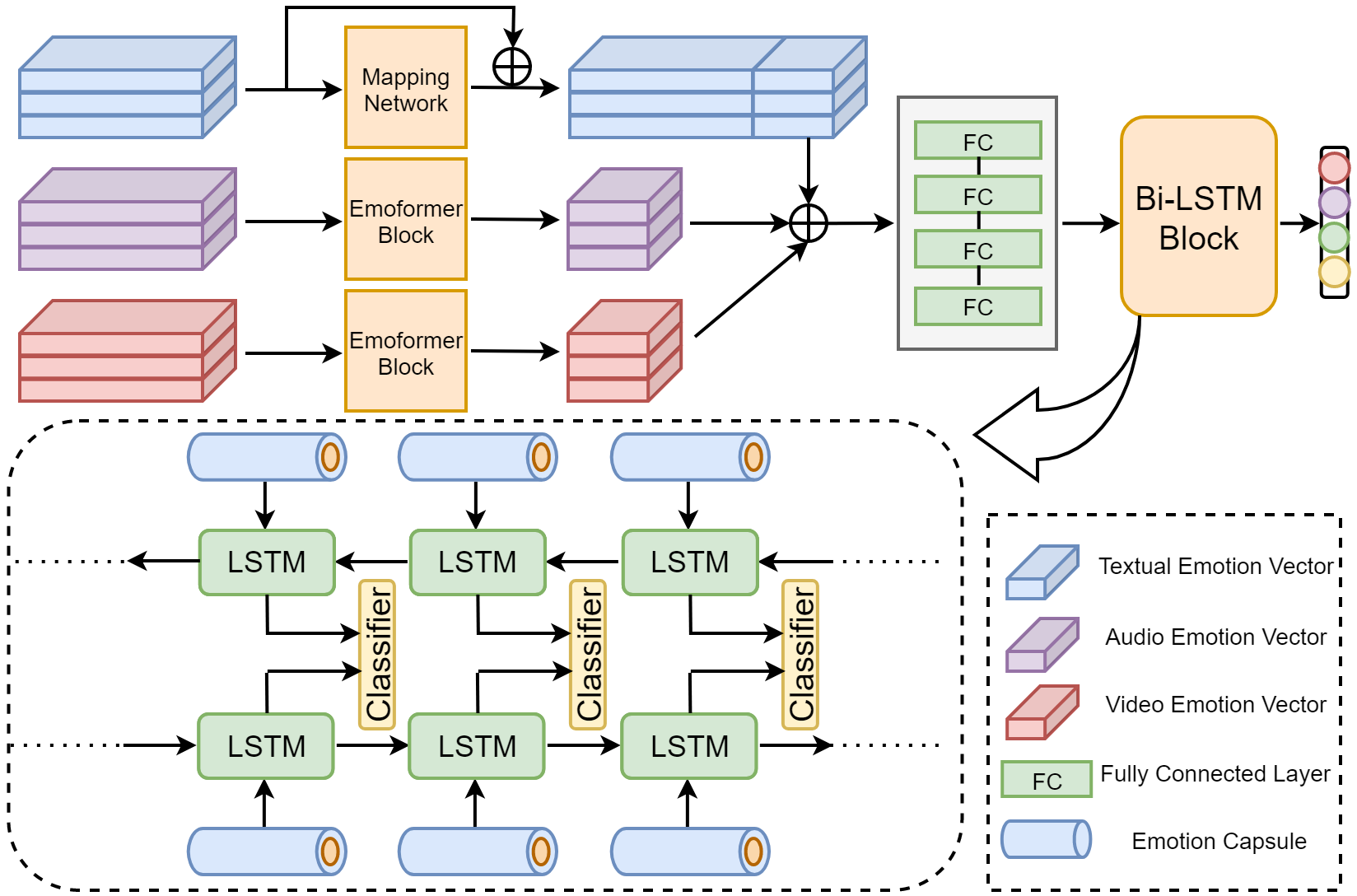} 
	\caption{Framework illustration of the EmoCaps based ERC model.}
	\label{fig3}
\end{figure*}
\subsection{Problem Definition}
\begin{table}[]
	\centering
	\begin{tabular}{c|c|c}
		\hline
		\multirow{2}{*}{\textbf{Parameter   Setting}} & \multicolumn{2}{c}{\textbf{Dataset}} \\ \cline{2-3} 
		& IEMOCAP            & MELD             \\ \hline
		Epochs                                        & 80                 & 80               \\
		Lr                                            & 0.0001             & 0.0001           \\
		Dr                                            & 0.1                & 0.1              \\
		Batch size                                    & 30                 & 30               \\
		Dim-T                                         & 100                & 600              \\
		Dim-V                                         & 256                & 256              \\
		Dim-A                                         & 100                & 300              \\ \hline
	\end{tabular}
	\caption{ Parameter settings detail of MELD dataset and IEMOCAP dataset. Epochs represents number of training epochs, Lr represents the learning rate, Dr represents the dropout rate. Dim-T represent the the total dimension of sentence vector and textual emotion vector, Dim-V and Dim-A represent the emotion vector dimensions of visual and audio modalities.}
\end{table}
Given a dialogue:{$u_1,u_2,u_3,…,u_n$}, where $n$ is the number of utterances. The purpose of conversation emotion recognition is to input a dialogue and identify the correct emotion classification of each sentence in the dialogue from the emotion label set $y$:{$y_1,y_2,y_3,...,y_m$}, where $m$ is the number of emotional label. 
\subsection{Unimodal Feature Extraction}

We extract the features of the utterance u, represented as U. In particular, when the input data is multi-modal, features of utterance U can be expressed as:
\begin{equation}
	U=[U_t,U_a,U_v]
\end{equation}	
where $U_t$ means textual feature, $U_a$ means audio feature, and $U_v$ means visual feature.

\noindent\textbf{Textual Feature Extraction}: In order to obtain good utterance representation, we use a pre-trained language model, BERT, to extract text feature vectors. BERT is a large general-purpose pre-trained language model proposed by Devlin et al. (2019), which can effectively represent the grammatical and semantic features of the utterance. Specifically, we first split the dialogue into a series of individual utterances, which are used as the input of the BERT-base model. Unlike other downstream tasks, we use the transformer structure to encode the utterances without classifying or decoding; then we get the sentence vector of every utterance with 512 dimensions. It should be noted that using a larger pre-tranined BERT model did not improve the performance, and a smaller BERT model couldn't get good enough performance.

\noindent\textbf{Audio Feature Extraction}: Identical to Hazarika et al. (2018), we use OpenSMILE (Eyben et al. 2010) for acoustic feature extraction. Specifically, in this work, we use the IS13 ComParE config file, which extracts a total of 6373 features for each utterance video, then we use the fully connected layer to reduce the dimensionality to 512 dimensions.

\noindent\textbf{Visual Feature Extraction}: We use the 3D-CNN model to extract video features, especially the facial expression features of the speaker. The 3D-CNN model can capture changes in the speaker's expression, which is important information in ERC. Specifically, we use 3D-CNN with three fully connected layers to get a 512-dimensional vector.

\subsection{Our Method} 
We assume that the emotion of the utterances in the dialogue depends on three factors:
\begin{itemize}
	\item The emotional tendency of the utterance itself.
	\item Emotional information contained in different modal of utterance.
	\item Context information
\end{itemize}
Based on the above three factors, our model EmoCaps is modeled as follows: We obtain three modal features of dialogue data: textual, audio, and visual; and input them into the Emoformer structure, then get the emotion vector of three modals and fuse them with sentence vector; finally, the context analysis model is used to obtain the emotion recognition result. The framework of the model is shown in Figure 4. It is worth noting that our text features, i.e., the sentence vector, are encoded by a transformer-based pre-trained language model, so we no longer use the self-attention mechanism but directly employ a mapping network to extract the emotion vector, then the residual structure concats sentence vector with emotion vector.

\noindent\textbf{Emoformer Block} Existing methods mainly use CNN, TextCNN, GRU, etc., to extract text feature vectors, which extract grammatical information weakly. At the same time, they only take the original feature vectors without emotional tendency as input. Based on this, we propose to use the Emoformer structure to extract the emotion vectors of various modalities. As shown in Figure 3, Emoformer has an Encoder structure similar to Transformer, but does not include the Decoder structure. A multi-head attention layer is used to get the emotional tendency feature from the original feature, both are connected through the residual structure, then emotion vector is obtained through a mapping network composed of 5 fully connected layers. The self-attention layer can be used to extract features that contain emotional tendencies or emotional factors effectively, and the residual structure ensures the integrity of the original information; finally the mapping network decouples features and reduces feature dimensions.

Identical to Vaswani et al. (2017), for a given input feature $U$, we calculate three matrix of query $Q$ $\in$ $\mathbb{R}^{{T}_{Q} \times {d}_{Q}}$, key $K$ $\in$ $\mathbb{R}^{{T}_{K} \times {d}_{K}}$ and value $V$ $\in$ $\mathbb{R}^{{T}_{V} \times {d}_{V}}$ by linear transformation from $U$:
\begin{equation}
	[Q,K,V]=U [W^Q,W^K,W^V]
\end{equation}
where ${T}_{Q}$, ${T}_{K}$, ${T}_{V}$ represent the sequence length of the $Q$, $K$, $V$, and ${d}_{Q}$, ${d}_{K}$, ${d}_{V}$ represent the dimensions of the $Q$, $K$, $V$, and $W^Q$ $\in$ $\mathbb{R}^{{d}_{Q} \times {d}_{m}}$, $W^K$ $\in$ $\mathbb{R}^{{d}_{K} \times {d}_{m}}$, $W^V$ $\in$ $\mathbb{R}^{{d}_{V} \times {d}_{m}}$.

Then we can express the formula of the self-attention layer as:
\begin{equation}
	A=softmax (\frac{QK^{T}}{\sqrt{d_k}})V
\end{equation}
where $A$ is the weight of value $V$, $d_k$ is equal to the dimension of $u$. In this way, multiple self-attention layers are concatenated to get Multi-Head Attention layer:
\begin{equation}
	MultiHead (A)=Concat (A_1,...,A_h) W
\end{equation}
where $A_1,...,A_h$ are the output of self-attention layers, $h$ is the number of layers, and $W$ is the weight parameter.

Then a residual connection with normalization layer is used to normalize the output of Multi-Head attention layer, and a Feed Forward layer is employed to get the output of the self-attention parts:
\begin{equation}	
	N=Norm (A+MultiHead (A))
\end{equation}
\begin{equation}
	F=max (0,NW_1+b_1)W_2+b_2\vspace{0.6ex}
\end{equation}
\begin{equation}
	G=Norm(F+MultiHead(F))
\end{equation}
where $W_1$, $W_2$ are the weight parameter, $b_1$, $b_2$ are the bias parameter.

Finally, the orignal features $U$ and the output of self-attention parts $G$ are connected through the residual structure, and a mapping network is employed to get the final output $E$:
\begin{equation}
	H=U \oplus G
\end{equation}
\begin{equation}
	E=Map (H)
\end{equation}
where the Map represents the mapping network, which consists of 5 fully connected layers.

Combine the above Eq. (2) to (9), we can get different modality emotion vectors from different input channels with Emoformer:
\begin{equation}
	[E_a,E_v,E_t]=Emoformer(U_a,U_v,U_t)
\end{equation}	
where $U_a, U_v, U_t$ represent the original input of audio,visual and textual features, and $E_a, E_v, E_t$ represent the emotion vectors of modalities. 

\noindent\textbf{Emotion Capsule} For the composition of the emotion capsule, we are based on the following rules: the text feature vector of the utterance contains grammatical and semantic features, emotion vector represents the emotional tendency of the utterance. Both are the main sources of conversation emotion recognition. Textual features most intuitively represent the meaning, emotions, characteristics, etc., of the utterance. However, visual features and audio features contain a few of emotional factors and emotional features, which can provide some emotional clues when the text features do not have sufficient emotional inclination. Therefore, sentence vector concats with three modalities' emotion vector to be an emotion capsule, which just like a "capsule", the emotion vector is "wrapped" by the sentence vector and "absorbed" by the context analysis model to determine the speaker's emotion finally. Our emotion capsule $O$ can be expressed as:
\begin{equation}
	O=U_t \oplus E_t \oplus E_v \oplus E_a
\end{equation}	
\begin{table*}[]
	\centering
	\begin{tabular}{c|c|c|c|c|c|c|c}
		\hline
		& \multicolumn{7}{c}{IEMOCAP}                                                                                         \\ \cline{2-8} 
		& Happy          & Sad            & Neutral        & Angry          & Excited        & Frustrated     & Average        \\ \hline
		BC-LSTM        & 43.40          & 69.82          & 55.84          & 61.80          & 59.33          & 60.20          & 59.19          \\
		CMN            & 30.38          & 62.41          & 52.39          & 59.83          & 60.25          & 60.69          & 56.13          \\
		DialogueRNN    & 33.18          & 78.80          & 59.21          & 65.28          & 71.86          & 58.91          & 62.75          \\
		DialogueGCN    & 42.75          & 84.54          & 63.54          & 64.19          & 63.08          & \textbf{66.99} & 64.18          \\
		DialogXL       & -              & -              & -              & -              & -              & -              & 65.95          \\
		DialogueCRN       & -              & -              & -              & -              & -              & -              & 66.20          \\
		DAG-ERC       & -              & -              & -              & -              & -              & -              & 68.03          \\
		MMGCN          & 42.34          & 78.67          & 61.73          & \textbf{69.00} & 74.33          & 62.32          & 66.22          \\ \hline
		EmoCaps & \textbf{71.91} & \textbf{85.06} & \textbf{64.48} & 68.99          & \textbf{78.41} & 66.76          & \textbf{71.77} \\ \hline
	\end{tabular}
	\caption{Experimental results (F1 score) on the IEMOCAP dataset. Average means weighted average. Some of the models only provide overall average results without results under each emotion category, so some data cells are lacking. }
\end{table*}
\noindent\textbf{Context Modeling} Since the same emotion has different expressions, and the same expression can express different emotions in different contexts, it is very difficult to infer the true emotions from a single word (Barrett, 2017). According to Grice's theory of implicature (1957), the meaning of a sentence can be canceled, so it is necessary to integrate context to infer the true meaning of a sentence. Therefore, contextual information is an indispensable part of conversation emotion recognition. Context information is divided into two parts: the information obtained from the previous moment is named emotional clue traceability, and the information obtained from the next moment is named emotional reasoning.

In this paper, we employ a Bi-directional LSTM model as the context analysis model to extract contextual information. 
In a conversation, we form a batch of emotion capsules of all utterances into the Bi-LSTM model in the order of dialogue, and each LSTM cell corresponds to an emotion capsule.  For the time i, in the forward propagation sequence, the contextual information $C_i$ at this moment is composed of the hidden state output of the LSTM cells at all previous moments, that is, the emotional clue traceability; in the backpropagation sequence, the contextual information at this moment is composed of the hidden state output of the LSTM cells at all following moments, which is emotional reasoning. The two fed into an MLP with fully connected layers and get the values of the utterance $u_i$ under each emotion label:

\begin{equation}
	l_i=ReLU(W_l C_i+b_l)
\end{equation}	
\begin{equation}
	P_i=softmax(W_{smax} l_i+b_{smax})
\end{equation}
where $W_l$, $W_{smax}$ are the weight parameter, $b_l$, $b_{smax}$ are the bias parameter.

Finally we choose the max value as the emotion label $y$ for the i-th utterance:
\begin{equation}
	y_i=\mathop{\arg\max}_{m}(P_i [m])\vspace{0.5ex}
\end{equation}

\section{Experiment Setting}

\subsection{Dataset}
\noindent\textbf{IEMOCAP} (Busso et al. 2008): The IEMOCAP dataset includes video data of impromptu performances or scripted scenes of about 10 actors. There are in total 7433 utterances and 151 dialogues in IEMOCAP dataset. At the same time, it contains audio and text transcription to meet the needs of multimodal data. In this data set, multiple commentators set the emotional labels of the utterances into six categories: including happy, sad, neutral, angry, excited and frustrated.

\noindent\textbf{MELD} (Poria et al. 2019): The MELD dataset contains 13708 utterances and 1433 conversations, which making up from TV series "Friends". It is also a multi-modal dataset containing video, audio, and text formats. In this dataset, multiple commentators set the emotional labels of the words into seven categories: including neutral, surprise, fear, sadness, joy, disgust, and angry.
\begin{table*}[]
	\centering
	\begin{tabular}{c|c|c|c|c|c|c|c|c}
		\hline
		& \multicolumn{8}{c}{MELD}                                                                                                           \\ \cline{2-9} 
		& Neutral        & Surprise       & Fear          & Sadness        & Joy            & Disgust       & Angry          & Average        \\ \hline
		BC-LSTM        & 73.80          & 47.70          & \textbf{5.40} & 25.10          & 51.30          & 5.20          & 38.40          & 55.90          \\
	
		DialogueRNN    & 73.50          & 49.40          & 1.20          & 23.80          & 50.70          & 1.70          & 41.50          & 57.03          \\
		DialogueGCN    & -              & -              & -             & -              & -              & -             & -              & 58.23          \\
		DialogXL       & -              & -              & -             & -              & -              & -             & -              & 62.41          \\
		DialogueCRN       & -              & -              & -             & -              & -              & -             & -              & 58.39          \\
		DAG-ERC       & -              & -              & -             & -              & -              & -             & -              & 63.65          \\
		MMGCN          & -              & -              & -             & -              & -              & -             & -              & 58.65          \\ \hline
		EmoCaps & \textbf{77.12} & \textbf{63.19} & 3.03          & \textbf{42.52} & \textbf{57.50} & \textbf{7.69} & \textbf{57.54} & \textbf{64.00} \\ \hline
	\end{tabular}
	\caption{Experimental results (F1 score) on the MELD dataset. Average means weighted average. The CMN model only for two-party conversation, but MELD is a multi-party conversation dataset. Some of the models only provide overall average results without results under each emotion category, so some data cells are lacking. }
\end{table*}
\subsection{Baseline Models}
\noindent\textbf{BC-LSTM} (Poria et al. 2017): Bc-LSTM uses Bi-directional LSTM structure to encode contextual semantic information, it does not recognize the speaker relationship.

\noindent\textbf{CMN} (Hazarika et al. 2018): It takes a multimodal approach comprising audio, visual and textual features with gated recurrent units to model past utterances of each speaker into memories.

\noindent\textbf{DialogueRNN} (Majumder et al. 2019): DialogueRNN uses different GRU units to obtain contextual information and speaker relationships. It is the first conversation emotion analysis model to distinguish between speakers.

\noindent\textbf{DialogueGCN} (Ghosal et al. 2019): DialogueGCN constructs a conversation into a graph, transforms the speech emotion classification problem into a node classification problem of the graph, and uses the graph convolutional neural network to classify the results.

\noindent\textbf{DialogXL} (Weizhou Shen et al. 2020): DialogXL use XLNet model for conversation emotion recognition to obtain longer-term contextual information.

\noindent\textbf{DialogueCRN} (Hu et al. 2021): DialogueCRN introduces the cognitive phase to extract and integrate emotional clues from context retrieved by the perceptive phase for context modeling.

\noindent\textbf{DAG-ERC} (Weizhou Shen et al. 2021): DAG-ERC  is a directed acyclic graph neural network for ERC, which provides a intuitive way to model the information flow between long-distance conversation background and nearby context.

\noindent\textbf{MMGCN} (Jingwen Hu et al. 2021): MMGCN uses GCN network to obtain contextual information, which can not only make use of multimodal dependencies effectively, but also leverage speaker information.
\begin{table}[]
	\centering
	\begin{tabular}{c|c|c}
		\hline
		\multirow{2}{*}{\textbf{Modality}} & \multicolumn{2}{c}{\textbf{Dataset}} \\ \cline{2-3} 
		& IEMOCAP           & MELD              \\ \hline
		Text                               & 69.49             & 63.51             \\
		Audio                              & 33.00             & 31.26             \\
		Video                              & 31.64             & 31.26             \\
		T+A                                & 71.39             & 63.73             \\
		T+V                                & 71.30             & 63.58             \\
		T+V+A                              & \textbf{71.77}    & \textbf{64.00}    \\ \hline
	\end{tabular}
	\caption{Performance (F1 score) of EmoCaps under different multimodal settings. T represent textual modality, A represent audio modality, and V represent visual modality.}
\end{table}
\subsection{Implementation}
For textual data, we use BERT model to obtain the sentence vector then get the textual emotion vector from a mapping network. For audio and visual data, we use Emoformer obtain the audio and visual emotion vector. 

As for the hyperparameter settings, we follow Li et al. (2021). For both of MELD dataset and IEMOCAP dataset, the epochs is set to 80, the learning rate is set to 0.0001, and the dropout rate is set to 0.1. The detailed parameter setting is shown in Table 1.

\section{Results and Analysis}
Our proposed model is compared with other state-of-the-art models on the IEMOCAP dataset and MELD dataset, which are under the same parameter conditions. The experimental results are shown in Table 2 and Table 3, our model has the best performance on both datasets.

\subsection{Compare with Other Baseline Models}
On the one hand, compared with existed methods, our model encodes sentences through a pre-trained language model to obtain a better utterance representation. On the other hand, our emotion capsule contains the emotional tendency of the utterance itself, combined with contextual information, can more effectively identify the speaker's emotion. The experimental results prove the rationality of our assumptions about the emotional factors in ERC.
\begin{table}[]
	\centering
	\begin{tabular}{c|c|c}
		\hline
		\multirow{2}{*}{Model} & \multicolumn{2}{c}{Dataset}    \\ \cline{2-3} 
		& IEMOCAP        & MELD           \\ \hline
		DialogueRNN            & 71.08          & \textbf{65.86} \\
		Bi-LSTM                & \textbf{71.77} & 64.00          \\ \hline
	\end{tabular}
	\caption{Performance(F1 score) under the multimodal setting of different models as context modeling model.}
\end{table}
\subsection{Various Modality}
Table 4 shows the performance of our model on the MELD dataset and the IEMOCAP dataset under different modality combinations. It is easy to find that the performance of multi-modal input is better than single-modal input. At the same time, among the three modalities of textual, audio, and visual, the textual modal has better performance than the other two modalities.

\subsection{Error Analysis}
As shown in Table 4, the performance of audio and visual modal is not good enough. For audio features, the frequency and amplitude of the sound features can only reflect the intensity of the speaker's emotions, not the specific emotional tendencies. Therefore, when certain emotions have similar frequencies and amplitudes, it is difficult to correctly distinguish the speaker's emotions only through audio data. For example, for two emotions of excited and fear, the frequency and amplitude characteristics in the audio mode are both at high values. Thus it's hard to distinguish the two emotions. For visual features, It is easy for us to judge the speaker's expression by facial features, but when the speaker hides his own expression, the video feature is not enough to judge the speaker's emotion. In addition, for a single video modality, the emotional changes in the context are unexplainable.

When the textual modality is added, the performance is significantly improved. In other words, textual modality play a major role in conversation emotion recognition, while audio and visual modalities can help improve the accuracy of recognition, which is consistent with the previous assumptions.
\begin{figure}[t]
	\centering
	\includegraphics[width=0.8\columnwidth]{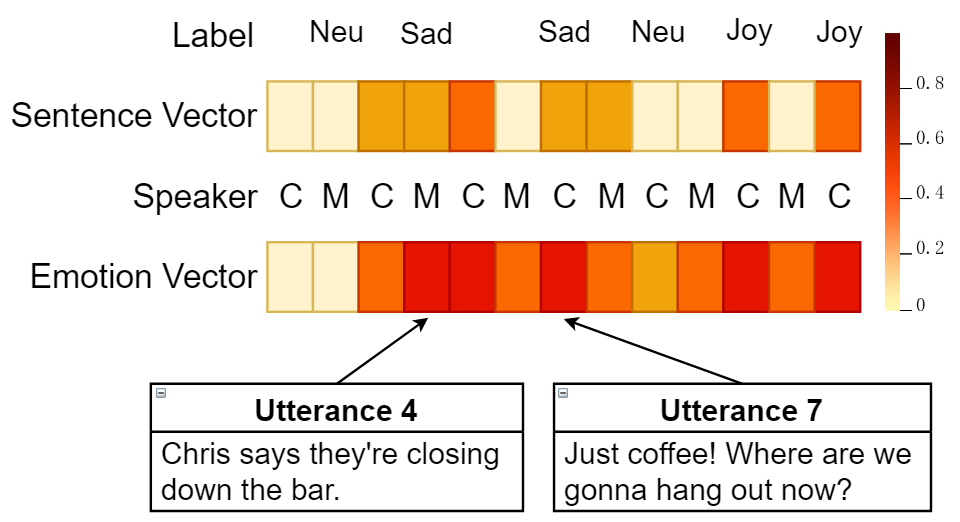} 
	\caption{Visualization of the heatmap in the "Friends" Season 3 Dialogue 2. Speaker C refer to Chandler and Speaker M refer to Monica.}
	\label{fig4}
\end{figure}

\subsection{Impact of Speaker Embedding}
In order to analyze the impact of speaker modeling on conversation emotion recognition, we use a variant of DialogueRNN as a context modeling model to test its performance on two benchmark datasets. As shown in Table 5, the performance of the DialogueRNN-based model on the MELD dataset is better than the LSTM-based model. The reason is that most of the MELD dataset belongs to multi-person dialogue situations, so the speaker modeling model (DialogueRNN-based) is more effective in identifying speaker emotions than the model not using speaker modeling (LSTM-based). However, in the IEMOCAP dataset, which is based on two-person dialogue situations, speaker modeling becomes insignificant.

Furthermore, compared with the LSTM-based model, using DialogueRNN-based or other models that include speaker modeling structures consumes more computing resources and time.

\subsection{Case Study}
Figure 5 shows the influence of emotion vector when emotion reversal in a conversation. At the beginning of the conversation, both speakers are in a neutral emotional state, while utterance 4 changes the situation that speakers' emotion turn into surprise and sadness. The sentence vector doesn't "understand" the reason why emotion change, but the emotion vector contains a negative emotion tendency which easier to get the correct emotion label. Utterance 7 shows that when the context is in the sad emotion, the emotion vector makes the utterance "biased" to "sad", while the sentence vector is in a neutral emotion. It proves the role of emotion vector in ERC.
\section{Conclusion}
In this paper, we propose a new multi-modal feature extraction structure, Emoformer, which is based on the transformer structure. Further, we design a new ERC model, namely EmoCaps. First, we use Emoformer structures to extract the emotion vectors of textual, audio, and visual modalities, then fuse the three modalities emotion vectors and sentence vectors to be an emotion capsule; finally, we employ a context analysis model to get the emotion recognition result. We conduct comparative experiments on two benchmark datasets. The experimental results show that our model performs better than the existing state-of-the-art models. The experimental results also verified the rationality of our hypothesis.
\bibliography{anthology,custom}
\bibliographystyle{acl_natbib}

\end{document}